\documentclass[pdflatex,sn-mathphys-num]{sn-jnl}

\usepackage{graphicx}%
\usepackage{braket}%
\usepackage{multirow}%
\usepackage{amsmath,amssymb,amsfonts}%
\usepackage{amsthm}%
\usepackage{mathrsfs}%
\usepackage[title]{appendix}%
\usepackage{xcolor}%
\usepackage{textcomp}%
\usepackage{manyfoot}%
\usepackage{booktabs}%
\usepackage{algorithm}%
\usepackage{algorithmicx}%
\usepackage{algpseudocode}%
\usepackage{listings}%

\theoremstyle{thmstyleone}%
\newtheorem{theorem}{Theorem}
\theoremstyle{thmstyletwo}%

\theoremstyle{thmstylethree}%
\newtheorem{definition}{Definition}%

\begin{document}

\title[Article Title]{
  
  \textbf{Advances in Machine Learning}: \\
  \large Where Can Quantum Techniques Help?
  
}


\author*{\fnm{Samarth} \sur{Kashyap}}\email{samarthk@iisc.ac.in}

\author*{\fnm{Rohit} \sur{K Ramakrishnan}}\email{rohithkr@iisc.ac.in}

\author{\fnm{Kumari} \sur{Jyoti}}\email{kumarijyoti@iisc.ac.in}

\author{\fnm{Apoorva} \sur{D Patel}}\email{adpatel@iisc.ac.in}

\affil{\orgdiv{Centre for High Energy Physics}, \orgname{Indian Institute of Science},\\ \orgaddress{ \city{Bangalore}, \postcode{560012}, \state{Karnataka}, \country{India}}}


\abstract{Quantum Machine Learning (QML) represents a promising frontier at the intersection of quantum computing and artificial intelligence, aiming to leverage quantum computational advantages to enhance data-driven tasks. This review explores the potential of QML to address the computational bottlenecks of classical machine learning, particularly in processing complex datasets.
We introduce the theoretical foundations of QML, including quantum data encoding, quantum learning theory and optimization techniques, while categorizing QML approaches based on data type and computational architecture. It is well-established that quantum computational advantages are problem-dependent, and so potentially useful directions for QML need to be systematically identified.
Key developments, such as Quantum Principal Component Analysis, quantum-enhanced sensing and applications in material science, are critically evaluated for their theoretical speed-ups and practical limitations. The challenges posed by Noisy Intermediate-Scale Quantum (NISQ) devices, including hardware noise, scalability constraints and data encoding overheads, are discussed in detail.
We also outline future directions, emphasizing the need for quantum-native algorithms, improved error correction, and realistic benchmarks to bridge the gap between theoretical promise and practical deployment. This comprehensive analysis underscores that while QML has significant potential for specific applications such as quantum chemistry and sensing, its broader utility in real-world scenarios remains contingent on overcoming technological and methodological hurdles.}

\keywords{Quantum Machine Learning, Quantum Computing, Classical Machine Learning, Quantum Data Encoding, Noisy Intermediate-Scale Quantum (NISQ), Quantum Principal Component Analysis, Quantum Neural Networks, Quantum Advantage, Quantum Sensing, Hybrid Quantum-Classical Models}



\maketitle

\section{Introduction}\label{sec1}

Machine learning has revolutionized the field of big data analysis, and it is an important component of the rapidly growing subject of artificial intelligence. The drive to process increasingly complex data sets necessitates rapidly growing computational resources. How to implement that efficiently has become a pressing and significant challenge. Quantum computational methods, with their ability to process information in ways that classical computers cannot, are being explored to find whether they can overcome some of these challenges and how.

The quantum formalism expands the space of variables that encode the data by allowing the possibility of superposition. The states are then described by the quantum density matrices that generalize the framework of classical probability distributions. This expansion may speed up certain tasks in sampling and data processing, through alternative techniques, and this improvement is labeled ``quantum advantage".
Of course, quantum computation is highly fragile against external disturbances, and the need to provide stable protected environment to quantum processors makes them expensive. The consequence is that quantum technology would be useful in areas where its advantage is large enough to offset its cost. The likely scenario for quantum technology is then a hybrid one: Special purpose quantum subroutines, carrying out problem-specific tasks, embedded in a larger classical computational setting. We want to identify such problem-specific quantum tasks and find their practical limitations. 

In Section \ref{ML}, we briefly introduce the classical machine learning theory, followed by a discussion on machine learning algorithms and their current applications. Next, in Section 3, we discuss the theoretical underpinnings of how quantum theory can influence computational models. Section 4 categorizes various quantum machine learning (QML) approaches based on data type and computational architecture, setting the foundation for an in-depth discussion of the proposed techniques in Section 5. We then outline the major challenges in QML research in Section 6, particularly in the context of Noisy Intermediate Scale Quantum (NISQ) devices. Section 7 highlights potential future directions, in quantum algorithms, their optimization techniques and various neural network architectures, and we conclude with a critical assessment of the prospects of QML in Section 8.

\section{Machine learning}\label{ML}

Over the past decade, machine learning (ML) has quickly become one of the popular Internet resources, with applications ranging from LLMs (such as GPT-4) to online advertisement optimization and medical diagnosis. As a field of research, it is concerned with developing and studying algorithms that can be automatically improved through experience using available data.
These algorithms build a \textit{model} based on a set of \textit{training data} to make predictions and decisions for an unknown set of inputs. Each data point, whether from the training data or unseen inputs, must comprise the same attributes based on which the model makes predictions---these attributes are called \textit{features}.

ML algorithms can be \textit{supervised} or \textit{unsupervised}. In supervised algorithms, the training data also consist of their output values, such as labels in an image classification problem. In unsupervised algorithms, training data does not include any expected output, but analogies are supplied, such as clustering problems to perform targeted advertising by demographic. Furthermore, there are other kinds of learning, such as semi-supervised and reinforcement learning, that do not fall neatly into either of these two categories.

Before we delve into the specifics and applications of machine learning, we must understand why it works, how it can fail, and how to quantify its efficacy. This will help us compare classical and quantum machine learning despite fundamentally different algorithms. We will first consider supervised learning, since the theory for unsupervised and other forms of learning is not as unified and is more problem-dependent.

\subsection{Classical learning theory}\label{learning-theory}

A supervised learning algorithm has access to the following ingredients \cite{shalevschwartz}:

\begin{itemize}
    \item The \textit{label set} \(\mathcal{Y}\), which is the set of all possible labels. The label may refer to a discrete label, as in a classification problem, or a continuous output value, as in a regression problem. It is also the output space of the algorithm's predictor.
    \item The \textit{instance space} \(\mathcal{X}\), which is the set of all possible inputs, or \textit{instance points}, the algorithm may be asked to label. The instance space is spanned by the features, which are the properties of the data that are recorded while preparing an input.
    \item The \textit{training data}, or the training set \(S=\{(\textbf{x}_1, y_1),\ldots, (\textbf{x}_m, y_m)\}\in \mathcal{X}\times\mathcal{Y}\). It is generated by a probability distribution \(\mathcal{D}\) over \(\mathcal{X}\times\mathcal{Y}\) that is unknown to the learner.
    \item The hypothesis class \(\mathcal{H}\), which is a set of maps \(h:\mathcal{X}\rightarrow\mathcal{Y}\) to choose the algorithm's predictor from.
\end{itemize}

The algorithm outputs a \textit{hypothesis} or \textit{prediction rule} \(h\in\mathcal{H}\), which can be used to predict the output of new instance points. The notation \(A(S)=h\) the algorithm \(A\) generates the hypothesis \(h\), when it receives the training set \(S\).
Our choice of letting training data be sampled from a joint distribution \(\mathcal{D}\) makes this model \textit{agnostic}
to the \textit{realization assumption} that the training data is generated by a perfect predictor in \(\mathcal{H}\).
This is a stricter, more realistic model of learning. Some extensions do not make this assumption, allowing for probabilistic imperfections. We will proceed with the agnostic model.

\subsubsection{Loss functions} To verify the hypothesis' usefulness, we need a way to measure its success. Since we only have access to the instance space through the distribution \(\mathcal{D}\), we can take an expectation of our metric over \((x, y)\sim\mathcal{D}\). To evaluate hypotheses against each other; we need the metric to be real, and to be able to compare expectations in a meaningful way, we need the metric to be positive (without loss of generality). We call this metric a loss function,

\begin{equation}
    l:\mathcal{H\times X\times Y}\rightarrow \mathbb{R}^+
\end{equation}
\\
It is obvious that a simple way to qualify a hypothesis is to ensure that \(l(h, x, y)=0\) when \(h\) is a perfect predictor for \(y\) given \(x\), and to ensure that the loss function increases as the hypothesis' prediction diverges from \(y\).

\subsubsection{Risk functions} As described, we may only sample instance points according to the distribution \(\mathcal{D}\). It, therefore, makes sense to take the expectation of the loss over samples of \((x, y)\) from \(\mathcal{D}\). It is known as the \textit{risk function},

\begin{equation}
    L_\mathcal{D}(h)\equiv\mathop{\mathbb{E}}_{(x, y)\sim \mathcal{D}}\left[l(h, x, y)\right]
\end{equation}
\\
In practice, however, we are unable to sample directly from the distribution, and we rely on the particular training data \(S\), so we use the \textit{empirical risk} over \(S\),

\begin{equation}
    L_S(h)\equiv\frac{1}{m}\sum_{i=1}^m {l(h, x_i, y_i)}
\end{equation}
\\
Again, since \(S\) is assumed to be the only form of data available to the learner, the obvious way to achieve a good predictor is to minimize \(L_S(h)\). This learning paradigm is called \textit{empirical risk minimization},

\begin{equation}
    \text{ERM}_{\mathcal{H}}(S)\in \mathop{\text{argmin}}_{h\in\mathcal{H}}L_S(h)
\end{equation}
\\
The form of the loss function varies in general and is usually dictated by a balance between the problem at hand and the computational constraints of minimizing it. By and large, the term "loss function" can be used to mean empirical risk, especially outside learning theory contexts, when discussing specific algorithms and applications.

\subsubsection{Learnability}

We need to define certain applicable conditions and metrics to determine if a machine learning algorithm will work. One such definition is \textit{agnostically probably approximate correctness}, introduced by Valiant \cite{Valiant1984} and improved upon
since then \cite{Vapnik1999} \cite{shalevschwartz}.

\begin{definition}[Agnostic PAC learnability]
    A hypothesis class \(\mathcal{H}\) is considered agnostically \textit{probably approximately correct} learnable if there exists a function \(m_\mathcal{H}:(0,1)^2\rightarrow\mathbb{N}\) and a learning algorithm \(A\) such that: For every \(\epsilon, \delta \in(0,1)\) and every distribution \(\mathcal{D}\) over \(\mathcal{X\times Y}\), 

    \begin{equation}
        L_\mathcal{D}(A(S_m))\leq\min_{h'\in\mathcal{H}}L_D(h')+\epsilon ,
    \end{equation}
    with probability \(1-\delta\) for all \(m\geq m_\mathcal{H}(\epsilon, \delta)\) where \(S_m\) is a training set of \(m\) instance points independently generated from \(\mathcal{D}\).

\end{definition}

Here, we are demanding that the algorithm can, most of the time (with probability \(1-\delta\)) get close (within \(\epsilon\)) to the least loss in the hypothesis class.

Another essential metric is quantifying whether a sample is sufficient to approximate the empirical risk function.

\begin{definition}[\(\epsilon\)-representative sample]
    A training set \(S\subset \mathcal{X\times Y}\) is \(\epsilon\)-representative with respect to a hypothesis class \(\mathcal{H}\), loss function \(l\) and distribution \(\mathcal{D}\) if
    \begin{equation}
        \forall h\in \mathcal{H}, \left|L_S(h)-L_\mathcal{D}(h)\right|\leq \epsilon .
    \end{equation}
\end{definition}

With these definitions, it has been shown\cite{shalevschwartz} that, with finite \(\mathcal{H}\) and bounded \(l\) within \(\left[0, l_\text{max}\right]\), \(\mathcal{H}\) is agnostically PAC learnable under the ERM paradigm with complexity,

\begin{equation}
    m_\mathcal{H}(\epsilon, \delta)\leq\left\lceil{\frac{2l_{\text{max}^2}}{\epsilon^2}\log\left(\frac{2|\mathcal{H}|}{\delta}\right)}\right\rceil .
\end{equation}

While the restrictions on \(\mathcal{H}\) and \(l\) seem harsh, in practice, hypothesis classes are defined by parametrized functions, and these parameters are stored on a computer with a fixed floating-point representation. This makes the seemingly infinite hypothesis class finite and allows us to validate the ERM paradigm for machine learning.
For example, if a learner with a loss function bounded by \([0,1]\) has \(n\) 64-bit parameters, the size of our hypothesis class is at most \(2^{64n}\), which tells us that the learning complexity of such an algorithm is upper-bounded by

\begin{equation*}
    m_\mathcal{H}(\epsilon, \delta)\leq\left\lceil{\frac{2}{\epsilon^2}\left((64n+1)-\log{\delta}\right)}\right\rceil\approx \mathcal{O}\left(\frac{128n-\log{\delta^2}}{\epsilon^2}\right) .
\end{equation*}

\subsubsection{Choice of hypothesis class}
The choice of hypothesis class is very important to determine the algorithm's accuracy. If \(\mathcal{H}\) is too dense, we run the risk of \textit{overfitting} the data, leading to low empirical risk over training data but high error on previously unseen data. As an example, given \({x_i, y_i}\), a classifier can predict \(y_i\) for a previously seen \(x_i\) and just return \(y_0\) for any \(x\) not in the training set.

Therefore, we must restrict the hypothesis class based on the prior information about the task. These restrictions are termed \textit{inductive bias} and reflect the belief of the learner that a member of \(\mathcal{H}\) is a low-risk hypothesis for the task. In practical terms, it is based on our partial knowledge about the distribution \(\mathcal{D}\). However, restricting \(\mathcal{H}\) too harshly can lead to a hypothesis that performs poorly on the training and the unseen data.

\subsubsection{No free lunch}
It is immediately evident that the most complex parts of designing an algorithm would be setting the hypothesis class and the loss function, using our knowledge and empirical observations of the theoretical distribution \(\mathcal{D}\). The obvious question is, does a universal learner \(\mathcal{H}, l\) exist that performs well on any distribution \(\mathcal{D}\)? The no-free-lunch theorem \cite{Shalev-Shwartz2013} states otherwise.

\begin{theorem}[No-free-lunch] Let \(A\) and \(B\) be two learning algorithms with instance space \(\mathcal{X}\) and label set \(Y\). Then, overall distributions \(\mathcal{D}\in \Delta(\mathcal{X\times Y})\), A and B will have the same average risk given the same sample size,
    \begin{equation}
        \sum_\mathcal{D}L_\mathcal{D}(A(S_{\mathcal{D}, m}))=\sum_\mathcal{D}L_\mathcal{D}(B(S_{\mathcal{D}, m})) ,
    \end{equation}
    where \(S_{\mathcal{D}, m}\) indicates a sample set of \(m\) independent instance points drawn from \(\mathcal{X\times Y}\) by \(\mathcal{D}\).
\end{theorem}

In the same vein, we can ask, why not just use all functions \(\mathcal{H}=\{h:\mathcal{X\rightarrow Y}\}\) as the hypothesis class? We have no better option with zero prior knowledge of the distribution \(\mathcal{D}\). However, the no-free-lunch theorem can be used to show that this class is not PAC learnable.

\subsubsection{VC dimension}
 While we have shown that finite hypothesis classes with bounded loss functions are learnable under the ERM paradigm, it is essential to look at infinite-size hypothesis classes as well, in the context of both a theoretical proof of learnability without using features of current computers as well as to understand purely quantum algorithms whose parameters are not necessarily discrete. In this context, we add a few definitions. We consider binary classifiers, but similar theorems and metrics exist for multiclass classifiers \cite{Natarajan1989} and regression learners \cite{Anthony1999} as well.

\begin{definition}[Shattering]
    Given a hypothesis class \(\mathcal{H}\) of functions \(\{h:\mathcal{X\rightarrow Y}\}\), a set \(C\subset \mathcal{X}\) is \textbf{shattered} by \(\mathcal{H}\), if \(\mathcal{H}\) can realize any labeling on \(C\) with some predictor within it.
\end{definition}

\begin{definition}[VC-dimension]\cite{Vapnik1971}\\
    The \textbf{VC-dimension} of a hypothesis class \(\mathcal{H}\), \(\text{VCdim}(\mathcal{H})\), is the maximal size of a set \(C\subset \mathcal{X}\) that can be shattered by \(\mathcal{H}\).
\end{definition}

It has been shown \cite{Blumer1989} that a finite VC-dimension guarantees agnostic PAC learnability. with complexity

\begin{equation}
    m_\mathcal{H}(\epsilon, \delta) = \Theta\left(\frac{d+\log{1/\delta}}{\epsilon^2}\right) .
\end{equation}
\\
It is, therefore, the measure to evaluate learnability, especially of hypothesis classes of infinite size. 

\subsection{Machine learning in practice}

Now that we have a handle on proving learnability, we discuss the general concepts of machine learning in practice. This section aims not to give a detailed overview of the algorithms, but to illustrate concepts such as parametrized networks and stochastic gradient descent that can carry over into quantum algorithms.

Usually, a hypothesis class is predicted using a parametrized network, where the inductive bias is applied by choosing the form of the function. For example, a simple transformation from input to output for a regression problem may be given by \(y=ax+b\), with \(a\) and \(b\) optimized by the learner to minimize empirical risk.

So far, we have not discussed the actual mechanism of risk minimization. There are two main approaches: analytic and numerical. Suppose the loss function and the transformation effected by the whole network are simple enough. In that case, the system of equations can be solved for stationary points of the risk as a function of the parameters, and the global minimum of risk can be exactly found. It is not always so simple; however, risk as a function of the parameters is a composite function of the parameters, and neural networks often effect very complicated transformations.

Therefore, the more common way of risk minimization is numerical. This is done by randomly initializing the parameters and using various techniques to ``ride" the curve and arrive at a stationary point. The most straightforward technique is gradient descent---at any point, the derivatives of the risk can be evaluated concerning each of the parameters, and by changing each parameter in the direction of its largest derivative, we move closer to a stationary point.

This technique, however, is fraught with several drawbacks: the stationary point reached may not be a global minimum, descent may take a long time depending on how the parameters are initialized, and there are certain types of functions where the learner never truly settles on a stationary point. Modifications have been proposed to address these issues, but we will not investigate their details.

\section{A theoretical formulation of quantum machine learning}\label{quantum-math}

In the classical learning theory discussed in Section \ref{learning-theory}, the training data is assumed to be sampled i.i.d from a distribution \(\mathcal{D}\) over \(\mathcal{X\times Y}\). When creating a quantum learner, there are two paths one can take for the data---the data can be sampled classically and encoded into quantum states, or the data can be sampled quantumly. The former is commonly attempted. However, the latter is feasible with the increasing prevalence of quantum sensing.

This section discusses both these approaches and their implications for quantum learning. We also introduce the mathematical framework behind each to provide a basis for the next section.

\subsection{Classical data and encoding}

With exact sampling as in classical learning theory, encoding the data into quantum states is a nontrivial task. Often, in quantum machine learning (QML), the biggest bottleneck for the algorithm runtime is the encoding of classical data.
It necessitates studying efficient state preparation methods and investigating the tradeoff between the encoding complexity and the circuit width.

An efficient classical machine learning runs in polynomial time with respect to the size of the data, quantified by the number of samples \(|S|=m\) and the number of features \(|\mathcal{X}|=n\).
Additionally, the precision of the features is usually decided by the computer architecture, which we can say is \(\tau\) bits.
Regarding QML, we may choose this freely; the number of qubits is not always the same as the number of features, with various encoding schemes available.
Schuld and Petruccione \cite{schuldbook} propose that an algorithm may be called \textit{qubit-efficient} or \textit{amplitude-efficient}, based on what we treat as the input of the algorithm---the number of qubits or the size of the Hilbert space.
The encoding efficiency of the two categories differs because the size of the Hilbert space is exponential in the number of qubits.

\subsubsection{Basis encoding}
Basis encoding is the simplest form of encoding, where we encode features as binary strings into a qubit set. For example, if we have a feature space \(\mathcal{X}=\{0,1\}^n\), we can encode each feature as a qubit, resulting in an n-qubit set.
It is apparent that basis encoding results in a wide circuit and is not very viable in the current state of quantum hardware.
The state preparation procedure for basis encoding is qubit-efficient, requiring at most \(n\tau\) gates. An advantage of this scheme is the ability to encode entire datasets into a single quantum state in superposition in a qubit-efficient manner, with \(\mathcal{O}(mn\tau)\) steps.
This technique is valuable in quantum random access memories (qRAM) and is an essential resource in many proposed quantum algorithms.

\subsubsection{Amplitude encoding}
In the amplitude encoding scheme, features are encoded as amplitudes of quantum states. A feature vector \(\textbf{x}\in\mathbb{C}^n\) can be encoded as

\begin{equation}
    \ket{\psi_\textbf{x}}=\frac{1}{\sum_i{\left|x_i\right|^2}}\sum_{i=1}^n x_i\ket{i} ,
\end{equation}
\\
where \(\ket{i}\) is the \(i^{\rm th}\) computational basis state of a \(\lceil\log_2{n}\rceil\)-qubit set. Entire datasets can also be encoded in this manner as

\begin{equation}
    \ket{\psi_S}=\frac{1}{\sqrt{m}}\sum_{k=1}^m\ket{\psi_{\textbf{x}^k}}\ket{k} ,
\end{equation}
\\
where \(\textbf{x}^k\) is the \(k^{\rm th}\) feature vector in the dataset. While amplitude encoding is the optimal encoding scheme to minimize circuit width, the state preparation procedure involves preparing an arbitrary state \(\sum_i x_i\ket{i}\).
This problem has been the subject of intense research and is lower-bounded by \(2^n/n\) two-qubit gates \cite{Plesch2011}, with most known algorithms performing slightly worse. These encoding schemes are amplitude-efficient but not qubit-efficient.
This property is a major bottleneck in near-term applications, as the amplitude encoding scheme has a circuit width low enough to be viable on modest quantum hardware. Still, the circuit complexity cost it introduces creates problems with gate noise, generally making the algorithm less efficient and reliable.

There have been attempts to use amplitude encoding in a qubit-efficient manner for specific kinds of data, such as
when the feature vectors are sparse \cite{gonzales2024}, or when the feature vectors are almost uniform \cite{schuldbook}.

\subsubsection{Other encoding schemes}
Other proposed encoding schemes have been summarized in various reviews of encoding \cite{schuldbook} \cite{khan2024}. It is important to note that the choice of encoding scheme is crucial for the efficiency of a QML algorithm.
The choice of encoding scheme is often a trade-off between the circuit width and the circuit complexity, and the dataset's characteristics must be kept in mind while choosing the scheme.

\subsection{Quantum data}

While the most obvious way to develop QML algorithms is to develop quantum circuits that perform the same functions as classical ML algorithms, the problems with encoding described in the previous section have led to an increased focus on quantum data. Advances in quantum sensing and error correction have enabled the use of these techniques practically soon.

While classical data comes with classical machine learning models that serve as inspiration for quantum algorithms, quantum data does not have the same luxury.
Research in this aspect has been slower, and much of it consists of algorithms designed for classical data but with the encoding step removed.
These algorithms have shown promise in essential applications such as quantum error correction and detection of phase transitions \cite{Cong2019}.
It is an open question, however, as to what kind of algorithms are best suited for quantum data, and our struggles with the unintuitive nature of quantum algorithms imply that they are unlikely to be of structure similar to the classical algorithms.

Our understanding of learning theory suffices for applying quantum algorithms to classical data, but to do the same with quantum data, it is sensible to have a new model of the sampling method since the learner receives the data as a quantum state instead of a classical sample.
The quantum agnostic example oracle QAEX(\(\mathcal{D}\)) was proposed by Arunachalam and de Wolf \cite{Arunachalam2018}:

\begin{equation}
    \text{QAEX}(\mathcal{D})\ket{0^{|\mathcal{X}|}, 0^{|\mathcal{Y}|}} = \sum_{(x, y) \sim \mathcal{D}}\sqrt{\mathcal{D}(x, y)}\ket{x,y} .
\end{equation}
\\
It is based on the weaker quantum PAC example oracle QPEX proposed by Bshouty and Jackson \cite{Bshouty1999}.

\subsection{Quantum learning theory}

\subsubsection{Sample complexity}
With a framework for the sampling methods a quantum learner can use, we can now look at extensions of classical learning theory to quantum learning theory.
In the case of classical data, no changes are needed. In learning theory, the model is a black box with access to the items mentioned in Section \ref{learning-theory}, which outputs a hypothesis; the fact that the model is quantum is irrelevant.
In the case of quantum data, however, the model must be extended to include the quantum agnostic example oracle, and the learning algorithm must be able to interact with this oracle instead of classical examples.

Arunachalam and de Wolf \cite{Arunachalam2018} propose a quantum analogue of the agnostic PAC learner:

\begin{definition}[Quantum agnostic PAC learnability]
A hypothesis class \(\mathcal{H}\) is considered agnostically probably approximately correct quantum learnable, if there exists a function \(m_\mathcal{H}:(0,1)^2\rightarrow\mathbb{N}\) and a quantum learning algorithm \(A\) such that:
For every \(\epsilon, \delta \in(0,1)\) and every distribution \(\mathcal{D}\) over \(\mathcal{X\times Y}\),

\begin{equation}
L_\mathcal{D}(A(S_m))\leq\min_{h'\in\mathcal{H}}L_D(h')+\epsilon ,
\end{equation}
\\
with probability \(1-\delta\) for all \(m\geq m_\mathcal{H}(\epsilon, \delta)\), where \(S_m\) is a training set of \(m\) invocations to the quantum agnostic example oracle \(QAEX(\mathcal{D})\).
\end{definition}

It was shown that a finite VC dimension guarantees agnostic PAC learnability with complexity.

\begin{equation}
    m_\mathcal{H}(\epsilon, \delta) = \Theta\left(\frac{d+\log{1/\delta}}{\epsilon^2}\right) .
\end{equation}
\\
Since a QAEX oracle can be used to sample classical data by measuring it immediately to Since a QAEX oracle can be used to sample classical data by measuring it immediately to get a sample \((x, y)\) with probability \(\mathcal{D}(x, y)\), any quantum learnable hypothesis class is also classically learnable, with at best a sample complexity advantage of a factor of \(n\) for the quantum learner.

While these results indicate no significant advantage to using QML algorithms, it is essential to note that agnostic PAC learnability has its roots in classical learning theory and may not be the best way to evaluate quantum algorithms.
Under the exact learning model \cite{angluin1988}, it has been shown \cite{servedio2004} that quantum algorithms have a polynomial advantage over classical algorithms.
Gavinsky \cite{gavinsky2012} developed a new model of learning called \textit{predictive quantum}(PQ) learning and demonstrated a relational hypothesis class that is polynomial-time learnable in the PQ model but needs an exponential number of samples in the classical (and quantum) PAC model.

Furthermore, noise plays a vital role in practical sampling.
It has been shown \cite{Bshouty1999} that quantum PAC learning sample complexity for the problem of learning a disjunctive normal form (DNF) expression in the presence of noise increases only polynomially. In contrast, classically, the problem is known to be intractable.
Similarly, Cross et al. \cite{Cross2015} showed that the problem of learning \(n\)-bit parity functions under-sampling noise has certain cases where quantum learning complexity grows only logarithmically. In contrast, classical learning complexity for the measured noisy oracle is superpolynomial.

\subsubsection{Time complexity}
While the sample complexity of a quantum learner is essential, the time complexity is crucial too; a learner who requires only a polynomial number of samples but an exponential amount of time to process them into a model is impractical.
To say that a hypothesis class is \textit{efficiently} learnable, under any model of learning, the algorithm must be able to process the samples in time polynomial in \(\{|\mathcal{X}|, |\mathcal{Y}|, 1/\epsilon, 1/\delta\}\).

Servedio and Gortler \cite{servedio2004} showed efficient PAC quantum learnability of several hypothesis classes based on factoring employing Shor's algorithm.
They further showed that the problem of distinguishing between a truly random function and a pseudorandom function generated by a class of one-way functions that are not based on factoring is efficiently quantum learnable.
Importantly, this proof does not depend on whether a quantum computer's class of one-way functions is invertible in polynomial time.

Furthermore, by relaxing the requirement that the algorithm efficiently learns for all distributions \(\mathcal{D}\) and restricting ourselves to some specific types of distributions, it has been shown that quantum learners offer up to exponential advantage over classical learners \cite{Bshouty1999} \cite{atici2007}.

\subsection{QML from a learning theory perspective}

To summarize Sections \ref{learning-theory} and \ref{quantum-math}, QML doesn't simply mean a plug-and-play black box that can be used to solve any machine learning problem.
Some problems are better left to classical learners, while quantum learners can offer a significant advantage for some combinations of data and problems.
The main advantage of QML is a gain in time complexity rather than sample complexity, but again, there are specific situations where QML may provide a sample complexity advantage.
QML has also proven more resilient to sampling noise than its classical
counterpart. Learning theory presents a unique formalism for realistic expectations in the intersection of two fields dominated by popular science hype.

In classical data on a quantum system, encoding presents a significant obstacle in realizing gains in circuit width. It is needed, nevertheless \cite{schuldbook}, since encoding is the only nonlinear transformation undergone by the data in QML, apart from weak nonlinearity in the final measurement.

The two types of data, classical and quantum, face opposite issues in research.
While quantum data has been studied extensively from a theoretical perspective, the lack of practical quantum data has made verifying the results with experiments difficult. On the other hand, classical data has been used in many experiments on quantum hardware. Still, the lack of a separate theoretical framework for classical data on quantum hardware has made it difficult to interpret the possible advantages compared to classical algorithms.

\subsection{Optimization processes}

While we have discussed the potential of learners as a black box, with knowledge of \(\mathcal{X}, \mathcal{Y}, \mathcal{H}\), and access to an oracle that outputs a hypothesis \(h\in \mathcal{H}\), we have not discussed the details of how the learner arrives at a hypothesis.
Classically, the dominant optimisation methods in supervised learning are variants and improvements of \textit{gradient descent}, where the learner adjusts the parameters of a hypothesis class in the direction of the gradient of the empirical risk to arrive at the minimum eventually.
Beyond theoretical learning contexts, loss, cost, and risk are used interchangeably to refer to the empirical risk.
Various improvements of gradient descent have been proposed, with different objectives such as faster convergence or avoidance of local minima.
An important method to find loss gradients concerning the model's parameters is \textit{automatic differentiation}, which, in essence, is the chain rule.
With the loss function \(l(h, x, y)\), and the model \(y = h(x) = f(\theta, x)\), the partial derivatives of the loss concerning each parameter are:

\begin{equation}
    \frac{\partial l}{\partial \theta_i} = \frac{\partial l}{\partial f}\frac{\partial f}{\partial \theta_i} ,
\end{equation}
\\
where both \(\frac{\partial l}{\partial f}\) and \(\frac{\partial f}{\partial \theta_i}\) are easy to compute after implementing \(f\) and \(l\).

If we want to train a quantum algorithm, we can take two approaches.
One is similar to the classical one, using gradient descent to optimise the parameters of a parametrised quantum circuit (PQC).
The other is to use quantum optimisation algorithms to optimise the parameters of the PQC in a computation run entirely on a quantum processor.
The former is termed a \textit{hybrid quantum-classical} algorithm, and the latter a \textit{fully quantum} one.
Most research geared toward near-term quantum computing applications is in the former category due to the existing literature and expertise in classical optimisation tools. In contrast, fully quantum algorithms are less well-explored.
Still, there are a few proposals for quantum optimisation algorithms, mostly
in unsupervised \cite{esma2007} and reinforcement learning\cite{Dunjko2016}.

One of the widely used tools in optimising hybrid algorithms is the concept of gradients of quantum circuits. The parameter-shift rule \cite{Schuld2019} shows how to calculate the gradient of a quantum circuit with respect to its parameters, using only two calls of the circuit with shifted parameters.
It has been used as a powerful tool in developing QML libraries like PennyLane \cite{bergholm2022}, facilitating many avenues of research in QML.
Apart from the parameter-shift rule, techniques like finite difference methods have also been adapted to find gradients. Some other tentative proposals \cite{Chen2024} to execute gradient descent on a quantum processor.

Gradient-free methods, such as the Nelder-Mead method \cite{Nelder1965} and sequential minimal optimisation \cite{Nakanishi2020}, have also been used in QML, especially because early experiments with gradient descent on quantum circuits exhibited the barren plateau problem \cite{McClean2018}.
The barren plateau problem is a phenomenon where the gradient of a quantum circuit with respect to its parameters is close to zero across vast regions of the parameter space, making gradient descent difficult.

\section{Taxonomy of QML approaches}

\begin{figure}[h]
    \centering
     \includegraphics[width=0.75\linewidth]{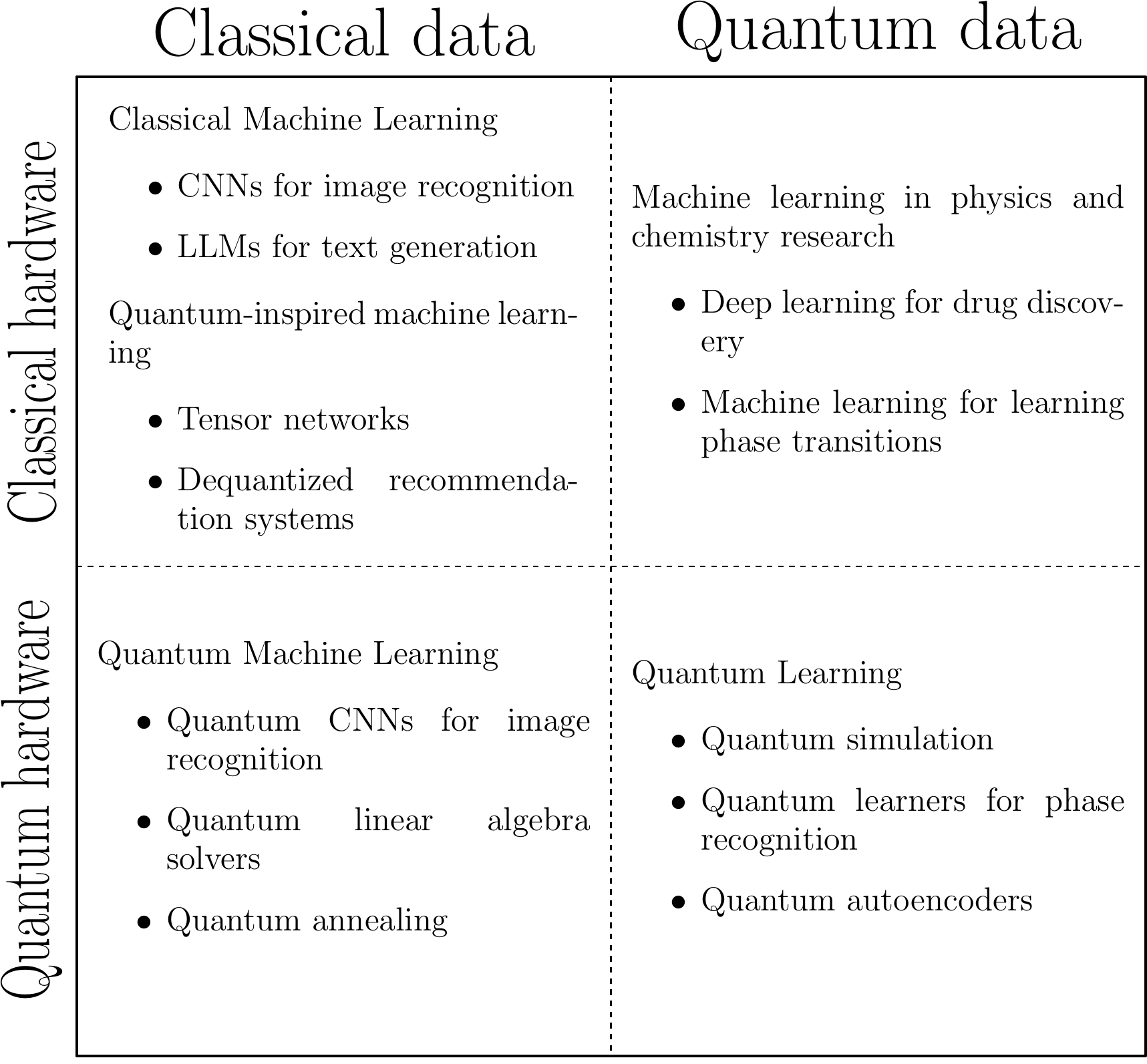}
    \caption{Classification of learners based on quantum/classical data and hardware}
    \label{fig:CQ_intersections}
\end{figure}

There are two ways to classify QML techniques based on the type of data used and the hardware used to run the algorithm. These are largely independent of each other, and various combinations have been proposed.
This choice is illustrated as a matrix of possibilities with the data and the device as the axes in \ref{fig:CQ_intersections}.
These categories, especially the latter, are not simply quantum/classical switches, and various hybrid modes exist.

In this Section, we discuss these techniques along with essential examples,
their observed advantages and disadvantages in practice, and their potential in light of the discussion on learning complexities in Section \ref{quantum-math}. In addition, we note a less-discussed, more abstract categorisation based on the inspiration behind the algorithms.

\subsection{Classical data on classical hardware}

While it may appear disingenuous to describe machine learning algorithms on classical hardware in a review of QML techniques, it is one of the most noteworthy practical contributions of QML so far---many classical ML algorithms in use are \textit{quantum-inspired}.
One example is the tensor network formulation, which was designed to model quantum many-body systems \cite{White1992} \cite{Penrose1971} \cite{Vidal2002}, but has found applications in traditional areas of machine learning \cite{Stoudenmire2016} \cite{Glasser2018}.
Other quantum-inspired algorithms appear in the field of linear algebra \cite{Chia2022} \cite{Chakhmakhchyan2017}.

Suppose a proposed quantum algorithm is better than the best-known classical algorithm. In that case, it has become a routine practice to attempt to develop a new classical algorithm inspired by the quantum one that would achieve the same speedup. This strategy has succeeded so often that it has given rise to the term \textit{dequantized} algorithms \cite{Arrazola2020}.
Dequantized algorithms have found applications in tasks like recommendation systems \cite{Tang2019}.

Although inspired by quantum algorithms, these algorithms are still classical and subject to the same efficiency and learnability restrictions as classical algorithms. They, however, play an essential role in advancing practical ML closer to established theoretical lower bounds.

\subsection{Quantum data on classical hardware}\label{quantum-data-classical-hardware}

Quantum data on classical hardware concerns using classical machine learning to aid traditional quantum computing. One such application is quantum state tomography \cite{Torlai2018}, which reconstructs entangled states by performing multiple measurements and is an essential tool in the validation and execution of many quantum algorithms.
Another application is the use of classical machine learning to detect phase transitions \cite{Carrasquilla2017} \cite{Wang2016} in many-body quantum systems, which has a potential use in the construction of quantum hardware.

This class of algorithms is an important area of research, as it combines the efficiency and practicality of classical machine learning with near-term quantum applications.

\subsection{Classical data on hybrid hardware}

This is the most explored approach to QML due to the familiarity of classical benchmark datasets and the relative ease of using a classical processor to train a parametrized quantum circuit (PQC).
In addition, since such QML algorithms are easy to implement on qubit simulators, there is a large body of work in this field, even though access to practical quantum hardware is limited.
Attempted scenarios include quantum convolutional neural networks (QCNNs) \cite{Kashyap2023}, convolutional neural networks \cite{Henderson2020}, recurrent quantum neural networks(QRNNs) \cite{Bausch2020}, and various quantum kernel methods \cite{havlicek2019}.

These proposals have shown promise in the time complexity of learning \cite{Hur2022} compared to classical counterparts. However, as seen in Section \ref{quantum-math}, better sample complexity is not guaranteed.
In practice, the time complexity advantage is often negated by the overhead of encoding classical data into quantum states or by a circuit width that is unrealizable in the near-term hardware.

These explorations have attracted considerable research due to their easy access to machine learning and quantum computing researchers. Still, they suffer from a lack of standardization in complexity analysis and performance measures.
Some recent works \cite{Liu2021} have rigorously demonstrated that QML algorithms can learn classical datasets that are not efficiently learnable by classical algorithms.
Most of these datasets, however, are constructed for the explicit purpose of demonstrating a quantum speed-up. It remains to be seen if any such algorithms can be helpful for real-world datasets in the near term.

\subsection{Quantum data on hybrid hardware}

This scenario has generated much interest due to the advantages of the hybrid approach mentioned above combined with the potential of near-term applications referred to in Section \ref{quantum-data-classical-hardware}.
Most research in this area considers the exact hybrid algorithms used for classical data minus the encoding step. Examples include QCNNs for phase detection in many-body systems and for quantum error correction \cite{Cong2019}, and quantum walks to solve the protein folding problem \cite{Casares2022}.

While these algorithms have demonstrated their prowess at performing tasks intractable for classical algorithms, most of them are adapted from hybrid algorithms for classical data, which are, in turn, adapted from classical algorithms.
These algorithms are a good starting point for classical data. Still, the lack of similarity between optimal quantum and classical (non-learning) algorithms for problems that have been studied extensively indicates that they may not provide the optimal way to handle quantum data.
We will discuss this point in Section \ref{inspiration-problem}.

In terms of complexity, this is the bare minimum scenario that can take advantage of the learnability advantages discussed in Section \ref{quantum-math}, and such advantages have been observed.
The sampling paradigm we described earlier is not necessarily adhered to in most quantum investigations, however, since the field of quantum sensing still has a long way to go before it can generate oracles such as QPEX and QAEX.

\subsection{Quantum hardware}

The topic of fully quantum machine learning algorithms has not been explored much. Some attempts include quantum Boltzmann machines \cite{Amin2018} and QAOA \cite{farhi2014}. They may be used with either classical or quantum data, but the quantum advantage over classical algorithms has not been established, and the investigations are in their infancy.
Although, as we pointed out in Section \ref{quantum-math}, there are potential gains in sample and time complexity that can be realized with fully quantum algorithms.

\subsection{The problem with extending classical ML to QML}\label{inspiration-problem}

Most developments in QML in recent years are extensions of classical paradigms into the quantum domain, such as QCNNs and QSVMs.
On the other hand, if we compare general quantum algorithms with a known advantage over their classical counterparts, for example, Shor's algorithm with the best-known classical factoring algorithms, we observe that the techniques are vastly different.
Much of the advantage of quantum algorithms comes from resources such as entanglement and superposition, which are absent in classical algorithms.
This naturally leads us to the question, if best quantum algorithms are so different from classical algorithms in general, is it appropriate to translate known classical ML algorithms into quantum ones?

A recent study \cite{bowles2024} considered various QML algorithms proposed recently and found that their performance did not change much when all entangling gates were removed from them. This implies that these algorithms are not taking advantage of quantum resources in principle.

Because good QML algorithms are not obligated to be intuitively understandable or similar to classical algorithms, just like known good quantum algorithms, the situation in which most QML research focuses on classical ML translated to a quantum setting is a significant cause for concern.
Furthermore, most of this QML research is benchmarked on classical datasets, such as MNIST, CIFAR-10 or housing price datasets. Though these datasets are valuable for benchmarking, there is no assurance that they represent the problems that will offer a quantum advantage.

There is a dire need to understand what aspects of a QML algorithm can help it learn better and to develop a framework that can be used to design algorithms that take advantage of these aspects.
This task is difficult since quantum computing is unintuitive, and explainable machine learning is in its infancy, even for classical algorithms.

\section{Some important QML developments}\label{sec:QML_approaches}

QML investigations have gained significant interest due to their potential to enhance computational efficiency in data-driven tasks. Despite theoretical promises, the practical feasibility of QML remains uncertain due to hardware limitations, noise, and data encoding challenges. This section explores notable QML applications, critically evaluating their advantages and limitations.

\subsection{Quantum Principal Component Analysis (QPCA)}

Quantum Principal Component Analysis (QPCA) is often cited as an example of how quantum speedups could be realized. Classical PCA reduces the dimensionality of the dataset by finding the eigenvectors of the data's covariance matrix and scales as \(\mathcal{O}(nd^2+d^3)\) where the dataset comprises of \(n\) \(d\)-dimensional features. Quantum PCA \cite{lloyd2014} deduces the properties of eigenvectors corresponding to the largest eigenvalues of a density matrix utilizing quantum phase estimation.

Quantum PCA can approximately predict the expectation values of observables of any given \(n\)-qubit state with a low-rank density matrix \(\rho\), with only \(\mathcal{O}(1)\) copies of \(\rho\). At the same time, a classical algorithm would need \(\Omega(2^{n/2})\) copies \cite{Huang2022}.
Crucially, this advantage is contingent on efficient quantum state preparation \cite{tang2021}. The overall speedup is negated if the dataset has to be classically encoded onto a quantum state. This process remains expensive, which leads us to believe that quantum PCA would be a valuable tool in settings with quantum sources of data or data that can be easily encoded.

\subsection{Quantum-enhanced sensing}

Quantum-enhanced sensing refers to using quantum entanglement and coherence to improve the precision of measurements in fields like gravitational wave detection, medical imaging, and atomic clocks. Quantum sensors have demonstrated a clear advantage in magnetometry, particularly in MRI applications, where QML techniques are employed for noise reduction and signal extraction \cite{Cerezo2021}.

Despite these advantages, quantum sensors are highly susceptible to decoherence.
While QML approaches, such as Quantum Variational Autoencoders (QVAEs), have been explored for denoising quantum sensor data, their practical implementation remains limited to small-scale demonstrations \cite{preskill2018}.
A significant challenge is the scalability of quantum-enhanced sensing techniques, as error rates increase significantly with system size.
Further research is required to determine whether QML-based sensing techniques can be extended beyond controlled laboratory conditions.

\subsection{Discrete logarithmic function dataset in QML}

The discrete logarithmic function dataset has been pointed out as a benchmark problem in QML due to its connection with Shor’s algorithm \cite{Arunachalam2018}. Quantum algorithms can estimate discrete logarithms efficiently using modular arithmetic, providing potential speed-ups in cryptographic analysis and number theory applications.

Still, the direct application of this dataset to machine learning tasks remains speculative. While quantum kernel methods have been proposed for discrete logarithm estimation, their performance advantage over classical algorithms has not been empirically validated.
Additionally, it is unclear what type of physical datasets would contain the discrete logarithm pattern.
As in other situations, the cost of encoding large datasets into quantum states remains a fundamental challenge \cite{weedbrook2012}.

\subsection{Material science and chemistry}

QML has the potential to simulate molecular interactions, optimise materials, and accelerate drug discovery. Hybrid classical-quantum Variational Quantum Eigensolvers (VQEs) have been designed to compute molecular ground states, reducing computational costs in quantum chemistry \cite{carleo2017}.
Near-term quantum devices still lack the qubit fidelity necessary for high-precision quantum chemistry calculations for modest-sized systems.
While quantum simulations are theoretically promising, classical approximations like density functional theory (DFT) remain more practical for most material science problems \cite{Cross2015}.
Additionally, integrating QML with high-throughput material screening methods is just beginning, and the added quantum advantage over existing classical methods is an open question.

\section{Challenges in QML}

While QML offers exciting theoretical insights, its practical realization remains uncertain. Most applications would require hardware, data encoding and scalability improvements before demonstrating tangible advantages over classical methods. Future work should focus on empirical benchmarks and real-world problem-solving rather than theoretical complexity reductions alone.
This Section explores the prominent obstacles impacting the future of QML.

\subsection{Noisy Intermediate-Scale Quantum (NISQ) era}

Translating the theoretical advantages of QML into practical applications is challenging, especially within the constraints of today's NISQ devices.
The NISQ era is defined by quantum processors containing up to 1,000 qubits, which have non-trivial capabilities but are still limited by noise and error correction challenges. These limitations hinder the development of QML algorithms that rely on quantum coherence for enhanced performance \cite{Heese2024}.

\subsubsection{Hardware limitations}
NISQ hardware faces issues related to stability, qubit fidelity and noisy gate operations, all of which affect the scalability and performance of QML.
For instance, Variational Quantum Algorithms (VQAs) depend on iterative computations susceptible to error accumulation, leading to decreased model accuracy. Hardware noise and limited gate fidelity restrict the circuit depth, making it challenging to implement deep learning operations on current quantum devices \cite{Cerezo2022}.

\subsubsection{Error rates and quantum noise}
Reducing quantum noise remains a significant hurdle in achieving reliable QML in the NISQ era. Errors arising from imperfect gate operations, decoherence and measurements introduce uncertainties, making it difficult to maintain reliable quantum states during extended computations. While techniques such as quantum error correction and quantum error mitigation are promising, they are not yet fully implementable on NISQ devices, resulting in reduced efficiency and accuracy of QML implementations on current hardware \cite{Cerezo2022}.

\subsubsection{Scalability constraints}
Without effective error correction, scaling QML models is severely constrained.
Building robust QML systems requires fault-tolerant quantum computers capable of supporting deep circuits necessary for intensive data processing.
As mentioned above, strategies are being tried, including quantum error codes and measurement error suppression. However, they remain challenging to implement on NISQ systems, and without them, scaling to larger system sizes is impossible \cite{Cerezo2022}.

\subsection{Classical-quantum interface and resource requirements}\label{sec:classical_quantum}

QML algorithms operate within a hybrid computational paradigm, combining the capabilities of classical and quantum systems to address computationally complex problems.
Several inherent challenges limit the practical applicability of this approach.
This Section examines the fundamental difficulties in integrating classical and quantum components and details theoretical versus practical resource constraints in QML implementations.

\subsubsection{Challenges in hybrid classical-quantum models}
Hybrid classical-quantum models are crucial in the NISQ era, enabling the utilization of possible quantum advantages while leveraging classical optimization techniques.
VQAs exemplify this hybrid approach, where quantum circuits perform computationally expensive tasks while classical components handle optimization.
Nevertheless, these models face several obstacles:

\begin{itemize}
\item \textbf{Barren plateaus:} QML models suffer from vanishing gradients in high-dimensional parameter spaces, making training inefficient.
The issue becomes more severe as system size increases, ultimately limiting scalability \cite{McClean2018}.
\item \textbf{Quantum noise and decoherence:} NISQ devices introduce errors due to gate noise, decoherence and imperfect qubit fidelity, which reduce the reliability of hybrid models. Even minor errors accumulate over iterations, making long training cycles impractical \cite{preskill2018}.
\item \textbf{Data encoding bottleneck:} Converting classical data into quantum states is resource-intensive; the cost of data encoding often negates any theoretical quantum speed-up. Even for algorithms with polynomial quantum advantages, data encodings require significant quantum memory and coherence time \cite{Huang2022}.
\end{itemize}

Recent experimental studies \cite{Huang2022} have demonstrated that while quantum-enhanced models outperform their classical counterparts in specific learning tasks, their efficiency depends on the ability to maintain coherence across multiple computational layers.
Their experiments with the Google Sycamore processor (40 qubits, 1300 quantum gates) validated quantum advantage in predicting physical system properties. Still, they also underscored the need for robust error mitigation strategies to make hybrid approaches viable.

\subsubsection{Resource requirements: Theoretical vs. practical constraints}
A key challenge in QML is the mismatch between theoretical predictions and practical resource availability. While quantum algorithms suggest computational advantages in machine learning tasks, their implementation on current hardware is constrained by several factors:

\begin{itemize}
\item \textbf{Qubit count and fidelity:} Theoretically, large-scale QML models require thousands of high-fidelity qubits to execute deep quantum circuits.
Today’s quantum hardware, however, is limited to noisy qubits with coherence times that restrict circuit depth \cite{Cross2015}.
\item \textbf{Error correction overheads:} Fault-tolerant quantum computing remains a distant goal. The surface code approach requires hundreds of physical qubits per logical qubit, making near-term error correction infeasible for large-scale QML applications \cite{Fowler2012}.
\item \textbf{Hybrid optimization complexity:} Many QML approaches use hybrid quantum-classical training loops. These require iterative updates between classical optimizers and quantum circuits, introducing latency and computational bottlenecks.
The overall runtime can exceed that of classical machine learning models when the data transfer overhead and noise correction requirements are considered \cite{Cerezo2021}.
\end{itemize}

\section{Future directions}\label{sec:future_directions}

While quantum-enhanced learning offers potential speed-ups in specific applications, its current feasibility is hindered by both hardware and algorithmic constraints.
Overcoming these limitations requires advances in quantum error correction, scalable qubit architectures, and efficient quantum data processing techniques.
Future work must focus on bridging the gap between theoretical quantum models and real-world computational feasibility to enable practical QML deployment \cite{preskill2018}.
This section outlines four significant areas where QML is expected to evolve.

\subsection{Quantum algorithms for machine learning}

Developing specialised quantum algorithms tailored for machine learning is essential to fully exploit quantum computational advantages. Current QML approaches often rely on adaptations of classical algorithms, limiting their efficiency on quantum hardware. Future research should focus on:

\begin{itemize}
\item \textbf{Specialized quantum hardware:} Advancements in superconducting qubits, photonic quantum processing, and trapped-ion systems that can significantly enhance QML performance by optimizing qubit connectivity and coherence times \cite{preskill2018, Arute2019}.
\item \textbf{Quantum feature engineering:} Designing quantum-specific representations of data that leverage entanglement and superposition to improve the learning performance \cite{Huang2022, Xia2018}.
\item \textbf{Algorithmic improvements:} Developing quantum-native learning models beyond classical paradigms, including quantum-enhanced kernel methods and quantum clustering algorithms \cite{Cerezo2021}.
\end{itemize}

\subsection{QML in quantum communication}

Quantum-enhanced communication protocols stand to benefit from QML techniques in areas such as secure information transfer. Key target areas in this domain include:

\begin{itemize}
\item \textbf{Error correction in quantum networks:} QML-driven optimization of quantum error correction codes to improve fidelity in long-distance quantum communication \cite{Fowler2012, Briegel1998}.
\item \textbf{Adaptive protocols:} Machine learning enhanced dynamic routing and noise prediction in quantum networks \cite{Lloyd2018}.
\item \textbf{Quantum cryptography:} Application of QML to cryptographic protocols, including quantum-secured authentication and secure multiparty computations \cite{Arunachalam2018}.
\end{itemize}

\subsection{Quantum optimization algorithms}

Quantum optimization algorithms are expected to play a significant role in solving high-complexity problems that classical optimization algorithms struggle with. Areas of development include:

\begin{itemize}
\item \textbf{Quantum annealing:} Exploring quantum annealers for combinatorial and discrete optimization tasks, with potential applications in logistics, finance, and material discovery \cite{kadowaki1998}.
\item \textbf{Variational quantum algorithms:} Enhancing quantum variational approaches to optimization problems by improving classical deep learning optimizers to mitigate barren plateaus \cite{McClean2018}.
\item \textbf{Quantum optimization for cryptographic applications:} Applying quantum optimization techniques to cryptographic problems, such as factorization and discrete logarithm calculations, for insights into security challenges and computational feasibility \cite{Arunachalam2018}.
\end{itemize}

\subsection{Quantum neural networks}

Quantum Neural Networks (QNNs) represent a promising avenue for merging quantum computing with artificial intelligence. Future research will likely focus on:

\begin{itemize}
\item \textbf{Hybrid classical-quantum neural networks:} Leveraging quantum circuits for feature extraction and embedding them within classical deep learning architectures \cite{Cong2019}.
\item \textbf{Quantum convolutional and recurrent models:} Adapting convolutional and recurrent neural network architectures to quantum circuits, improving pattern recognition in quantum-enhanced image processing and time-series analysis \cite{Schuld2019}.
\item \textbf{Scalability and trainability:} Addressing challenges in training QNNs efficiently, including noise mitigation, quantum gradient descent methods, and error resilience in quantum backpropagation \cite{benedetti2019}.
\end{itemize}

By addressing these future directions, QML can transition from theoretical explorations to practical deployment across various fields, from quantum-enhanced artificial intelligence to secure quantum communication networks.

\section{Outlook}

QML has emerged as an area of interest at the intersection of quantum computing and classical machine learning, driven by the possibility of computational speed-ups and new data processing paradigms. Despite the theoretical potential, however, there is little concrete evidence that QML can provide a consistent advantage over classical machine learning methods in real-world applications.
Although specific quantum algorithms demonstrate efficiency improvements in isolated scenarios, such as quantum-enhanced sensing and variational optimization methods, the extent to which these advantages generalize beyond some restricted examples remains an open question.

One of the fundamental challenges in QML is the high cost of data encoding, which often negates any theoretical speed-up promised by quantum algorithms.
In practical implementations, quantum-classical hybrid models have become the predominant approach, as purely quantum architectures remain infeasible due to hardware limitations, noise, and short coherence times.
Even for hybrid models, the empirical quantum advantage over classical machine learning remains unclear, particularly in the NISQ era, where the error rates and quantum resource constraints impose severe limitations on achievable circuit depth and problem complexity.

Furthermore, the lack of a well-defined framework for benchmarking QML against classical alternatives poses a significant issue. It is unclear whether QML will ever offer a meaningful advantage for practical data-driven applications or whether its benefits will be limited to specific tasks, such as investigations in quantum chemistry or condensed matter physics.
Most likely, practical QML applications would be in situations where the input data is quantum or at least collected in a form suitable for quantum processing.

While advances in quantum sensing and quantum communication may provide some near-term utility, the broader field of QML remains largely underdeveloped.
Future work should prioritize realistic benchmarks, practical feasibility studies, and a clearer understanding of the conditions under which quantum models can surpass their classical counterparts. 

\section{Acknowledgements}

The authors thankfully acknowledge support from the research project `Quantum Algorithms and Simulations' under DIA-RCoE at IISc.

\bibliography{sn-bibliography}

\end{document}